\documentclass[review]{elsarticle}

\usepackage{lineno,hyperref}
\modulolinenumbers[5]

\journal{Expert Systems}

\usepackage{graphicx}
\usepackage{amsmath}
\usepackage{amssymb}
\usepackage{booktabs}
\usepackage{threeparttable}

\usepackage{listings}

\lstdefinestyle{yaml}{
     basicstyle=\color{blue}\footnotesize,
     rulecolor=\color{black},
     string=[s]{'}{'},
     stringstyle=\color{blue},
     comment=[l]{:},
     commentstyle=\color{black},
     morecomment=[l]{-}
 }

\usepackage[final]{changes}

\definechangesauthor[color=red]{R1} 
\definechangesauthor[color=purple]{R2}
\newcommand{\rev}[1]{\added[id=R1]{#1}}
\newcommand{\del}[1]{\deleted[id=R1]{#1}}

\newcommand{\appropto}{\mathrel{\vcenter{
  \offinterlineskip\halign{\hfil$##$\cr
    \propto\cr\noalign{\kern2pt}\sim\cr\noalign{\kern-2pt}}}}}

\begin{document}

\begin{frontmatter}

\title{Efficient Adaptive Ensembling for Image Classification
}

\author{Antonio Bruno$^\dagger$, Davide Moroni$^{\dagger,*}$, Massimo Martinelli$^\dagger$}
\address{Institute of Information Science and Technologies (ISTI) \\ National Research Council of Italy (CNR)\\ \ead{antonio.bruno@isti.cnr.it, davide.moroni@isti.cnr.it, massimo.martinelli@isti.cnr.it}
Via Moruzzi 1, Pisa, Italy}
\fntext[myfootnote]{$^\dagger$.}

\ead[url]{www.isti.cnr.it}

\cortext[mycorrespondingauthor]{Corresponding author}
\address{$\dagger$ These authors have contributed equally to this work and share first authorship}


\begin{abstract}
In recent times, \del{except for} \rev{with the exception of} sporadic cases, the trend in Computer Vision is to achieve minor improvements \del{over} \rev{compared to } considerable increases in complexity. 
  To reverse this \del{tendency}\rev{trend}, we propose a novel method to boost image classification performances without \del{an increase in}\rev{increasing} complexity.
  To this end, we revisited \emph{ensembling}, a powerful approach, \del{not often adequately used due to its nature of increased complexity and training time, making it viable by specific design choices} \rev{often not used properly due to its more complex nature and the training time, so as to make it feasible through a specific design choice}. First, we \del{end-to-end} trained two EfficientNet-b0 \rev{end-to-end} models (known to be the architecture with 
  the 
  best overall accuracy/complexity trade-off \del{in} \rev{for} image classification) on disjoint subsets of data (i.e. bagging). Then, we made an efficient adaptive ensemble by performing fine-tuning of a trainable combination layer. In this way, we were able to outperform the state-of-the-art by an average of 0.5$\%$ on the accuracy\rev{,} with restrained complexity both in terms of the number of parameters (by 5-60 times), and the FLoating point Operations Per Second \rev{FLOPS}  \del{(} by 10-100 times \del{)} on several major benchmark datasets.
\end{abstract}

\begin{keyword}
Deep Learning \sep Ensemble \sep Convolutional Neural Networks \sep EfficientNet \sep Image Classification
\end{keyword}

\end{frontmatter}

\section{ Introduction }\label{intro}
\subsection{\del{Motivations}}
Computer vision is one of the fields that most benefit from deep learning, continuously  improving the state-of-the-art (SOTA) using Convolutional Neural Networks (CNNs) and Visual Transformers. In nearly all computer vision scenarios, complexity grows exponentially, even for minimal improvements, both in terms of the number of parameters and in FLoating point Operations Per Second (FLOPS). Table \ref{tab:evolution} briefly shows the evolution of the SOTA on the ImageNet classification task. It can be observed that the trend of improvements achieved only through high complexity growth was temporarily slowed down by the introduction of EfficientNet architecture (and in particular with EfficientNet-b0 attaining the best accuracy/complexity trade-off ) \cite{efficientnet}. \del{The previous} \rev{This} also applies to other image classification \rev{datasets} (e.g. CIFAR) and \rev{to} \del{C}\rev{c}omputer \del{V}\rev{v}ision tasks based on CNNs (e.g. object detection and segmentation).

\begin{table*}[h]
\caption{Evolution of the state-of-the-art on the ImageNet classification task: as can be seen, complexity in models having accuracy $>$ 80\%  (both in the number of parameters and FLOPs) grows exponentially \del{in spite of minimal} \rev{despite the slightest} improvement. The same trend can be noticed in other computer vision tasks. \emph{N.B. only some architectures providing relevant improvements are shown \del{in this table}.}}
\newcommand\T{\rule{0pt}{2.5ex}}
\begin{center}
\begin{tabular}{lcccc}
    {\bf Model} &  {\bf Year} & {\bf Accuracy} & {\bf Parameters}  & {\bf FLOPs}\\
    \hline
    \T
    AlexNet \cite{alexnet} & 2012 & 63.3\% &  $\approx$ 60M & $\approx$ 0.7G\\
    InceptionV3 \cite{inceptionv3} & 2015 & 78.8\% & $\approx$ 24M & $\approx$ 6G\\
    ResNeXt-101 64x4 \cite{resnetx} & 2016 & 80.9\% & $\approx$ 84M & $\approx$ 16G\\
    EfficientNet-b0 \cite{efficientnet} & 2019 & 77.1\% & $\approx$ 5.3M & $\approx$ 0.4G\\
    EfficientNet-b7 \cite{efficientnet} & 2019 & 84.3\% & $\approx$ 67M & $\approx$ 37G\\
    Swin-L \cite{swin} & 2021 & 87.3\% & $\approx$ 197M & $\approx$ 103G\\
    NFNet-F4+ \cite{nfnet} & 2021 & 89.2\% & $\approx$ 527M & $\approx$ 215G\\
    ViT-G/14 \cite{vit} & 2021 & 90.45\% & $\approx$ 1843M & $\approx$ 965G\\
    CoAtNet-7 \cite{coatnet} & 2021 & 90.88\% & $\approx$ 2440M & $\approx$ 2586G
\end{tabular}
\end{center}
\label{tab:evolution}
\end{table*}

\subsection{\del{Ensembling}}
\rev{Among the various machine learning approaches}, \emph{ensembling} is \del{the} \rev{a} technique that combines several models, called weak learners, in order to produce a model with better performance than any of the weak learners alone \cite{ensemble_orig}. Usually, the combination is accomplished by aggregating the output of the weak learners\rev{,} generally \rev{this is made} by voting (resp. averaging) for classification (resp. regression). Other aspects, such as ensemble size (i.e. number of weak learners) and ensemble techniques (e.g. bagging, boosting, stacking), are crucial for obtaining a satisfactory result. Since it requires the training of several models, ensembles make\rev{s} the overall validation much more expensive, and model complexity grows at least linearly \del{with respect} \rev{compared} to the ensemble size\del{;}\rev{.} 
\del{m}\rev{M}oreover, ensembling is a time-consuming process, and this is the main reason preventing a more extended use in practice, especially in \del{C}\rev{c}omputer \del{V}\rev{v}ision. \del{By converse}\rev{On the contrary}, this work shows \rev{that} our technique \del{to exploit} \rev{exploits} this powerful tool with \del{restrained}\rev{limited} resources (e.g. \del{with respect}\rev{compared} to \rev{the} model complexity, validation time and training time).

\subsection{\del{Content outline}}
This work shows how applying a well-defined ensembling strategy\rev{,} using an efficient basic model as the core\rev{,} can improve the state-of-the-art in \del{C}\rev{c}omputer \del{v}\rev{v}ision tasks,  preserving a competitive performance/complexity trade-off.
In Section  \del{\ref{sec:related_work} we survey related works why in Section} \ref{sec:our_work} we describe our design\del{ing} strategy in detail (e.g. model, ensembling strategy, validation), focusing on \del{introducing}\rev{ on the introduction of} the main novel aspects. Experimental results and data description are shown in  Section \ref{sec:results}, while an exhaustive discussion is provided in the last section.

\section{\rev{Related Work}}\label{sec:related_work}
\rev{In recent  years, the demand for  intelligent systems based on image processing has also grown on the push of emerging business markets. In this context, the capacity to deal with large-scale collections of images has not only to face significant technological challenges but must be shown to be cost-effective and, ultimately, sustainable. Indeed, the carbon impact of artificial intelligence (AI) is a concern that has been well recognized \cite{dhar2020carbon}, favouring the adoption of green AI paradigms  \cite{schwartz2020green}. In particular, in order to reduce the carbon footprint of AI and make it cost-effective in new markets, it is possible to follow several pathways, including decentralized approaches based on federated learning (therefore not requiring energy-consuming data transfer) \cite{bonawitz2019towards} or devising \emph{ad hoc} low-consumption hardware specific for modern deep learning algorithms  \cite{sze2017hardware}. 
Other methods deal with the AI model itself, proposing its simplification or optimization; well-known techniques, mainly suited for inference, include parameter quantization and pruning, compressed convolutional filters and matrix factorization, network architecture search, and knowledge distillation \cite{goel2020survey}. In this paper, instead, we propose a method for achieving greener models both in training and inference by resorting to ensembling.}

\rev{Ensembling consists in a machine learning approach in which a set of \emph{weak learners} (or \emph{basic models}) is turned into a \emph{strong learner} (or \emph{ensemble model})\cite{ensemble_orig, ensemble_survey2}. The set of weak learners might consist of homogenous models (i.e., they are all from the same family or architecture) or might be heterogeneous, i.e., the basic models belong to different machine learning paradigms. The basic example is to put together multiple models trained for solving the same classification or regression task and then combine them in some fashion, e.g., by performing majority voting in the case of classification or averaging in the case of regression. The scope of performing Ensembling is generally related to the desire to reduce the bias or variance that affects a machine learning task \cite{dong2020survey}.}
\rev{As it is well known, a low-complexity model might have a significant error in attaining adequate performance on a dataset, even during training. This is commonly due to the low representation capabilities of simple models that can only capture some of the complex patterns in the training datasets. Such error during training is referred to as the bias of the model. By converse, very complex models have many degrees of freedom to completely stick to the training dataset and convey excellent performance during training. However, they apprehend not only the relevant features of the problem but also learn unimportant features of the training dataset. This results in relatively inadequate performance during test and validation: the model needs to be more balanced to the training dataset and reach good general results, having scarce generalization capabilities. Such an issue is usually indicated as a high variance of the model. }
\rev{The three primary techniques for conducting ensembling are bagging, boosting, and stacking.
In general, bagging decreases the variance among the weak classifiers, while boosting-based ensembles reduce bias and variance. Stacking is generally employed as a bias-reducing procedure.
In more detail, the bagging technique involves partitioning the training datasets into distinct subsets based on specific criteria, such as equalizing class distributions within each subset. Subsequently, a weak classifier is trained using each subset of the training set. Ideally, these classifiers possess low bias on the training set but may exhibit high variance. The outputs of these individual classifiers are then combined through weighted voting or a weighted average using a specially designed layer. This fusion of weak classifiers forms the strong classifier, which tends to have reduced variance. It's worth noting that the weak classifiers can be trained independently and in parallel. 
In boosting instead, weak classifiers are very simple and low complexity but are trained cleverly, for example, using cascading.}
\rev{Ultimately, stacking commonly involves the consideration of diverse weak learners with varying characteristics. The training process takes place concurrently, and a final amalgamation is achieved by training a meta-model that generates predictions based on the collective inputs from the various weak models.}
\rev{In general, all of these approaches have been used in conjunction with deep learning models. The review \cite{ganaie2021ensemble} presents some recent literature on the subject systematically.}
\rev{ In this paper, we propose using bagging  in an original way that allows us to obtain superior results with respect to the state of the art while decreasing the computational burden.}

\section{Efficient Adaptive Ensembling}\label{sec:our_work}
\subsection{Efficiency}
\begin{figure*}[h]
\centering
\resizebox{.9\textwidth}{!}
{\includegraphics{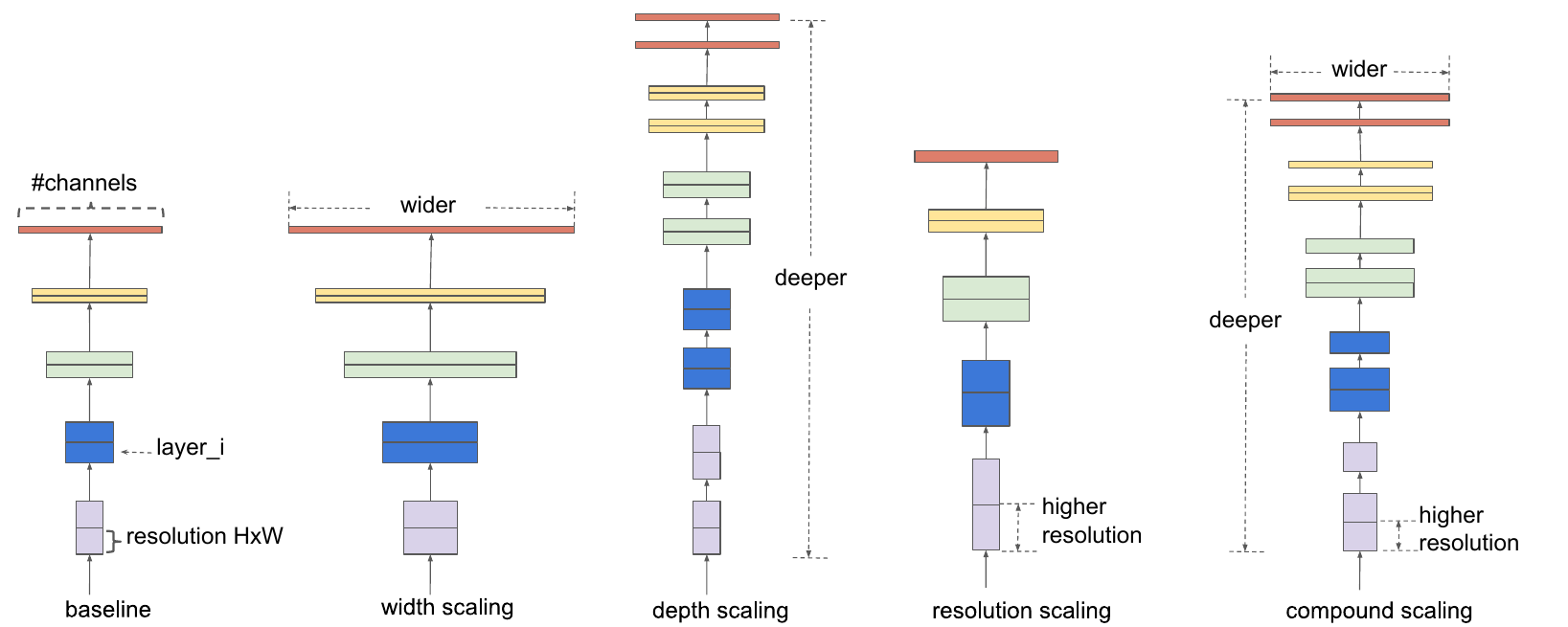}}
\caption{Example of scaling types, from left to right: a baseline network example, conventional scaling methods that only increase one network dimension (width, depth, resolution) and, at the end, the EfficientNet compound scaling method that uniformly scales all three dimensions with a fixed ratio. Image taken from the original paper \cite{efficientnet}.}
\label{fig:scaling}
\end{figure*}

At the foundations of the efficiency of the proposed method lies the basic core model adopted in this work: EfficientNet \cite{efficientnet}. As the name suggests, EfficientNet improves the classification quality with lower complexity compared to models having similar classification performances. This is possible since EfficientNet performs optimised network scaling, given a predefined complexity. As shown in Figure \ref{fig:scaling}, in the CNN literature, there are three main types of scaling: \rev{ \emph{depth scaling,width scaling and input scaling}.
Depth scaling consists in increasing the number of layers in the CNN; it is the most popular scaling method in the literature and allows detecting  features at multiple levels of abstraction.
Width scaling consists in increasing the number of convolutional kernels and parameters or channels, giving the model the capability to represent different features at the same level.
Input scaling is represented by the increase in size/resolution of the input images, which allows \rev{for} capturing additional details.
}

Each of these scalings can be manually set or \del{by}\rev{via} a grid search. However, they increase the model complexity, usually exponentially, with tons of new parameters to tune and, after a certain level, scaling appears not to improve performances.
The scaling method introduced in \cite{efficientnet} is named \emph{compound scaling}. It suggests that \del{strategically performing} \rev{the strategic execution of} all scaling together \del{delivers}\rev{provides} better results because it is argued that they are dependent. Intuitively, they introduce the \emph{compound coefficient} $\phi$ representing the total amount of resources available to the model and find the optimal scaling combination given such a constraint, following the rules in Equation  \ref{eq:compoundscaling}. In this way, the total complexity of the network is approximately proportional to $2^{\phi}$ (see the original paper for more details). 
\begin{equation}
\begin{gathered}
\text{depth: } d = \alpha^{\phi}  \quad   \text{width: } w = \beta^{\phi}   \quad   \text{resolution: } r = \gamma^{\phi} \\
\text{such that} \quad \alpha  \cdot \beta^2 \cdot \gamma^2 \approx 2 \quad \text{and} \quad \alpha \geq 1, \beta \geq 1, \gamma \geq 1
  \label{eq:compoundscaling}
 \end{gathered}
\end{equation}

\subsection{Adaptivity}
\begin{figure*}[h]
\centering
\resizebox{.6\textwidth}{!}
{\includegraphics{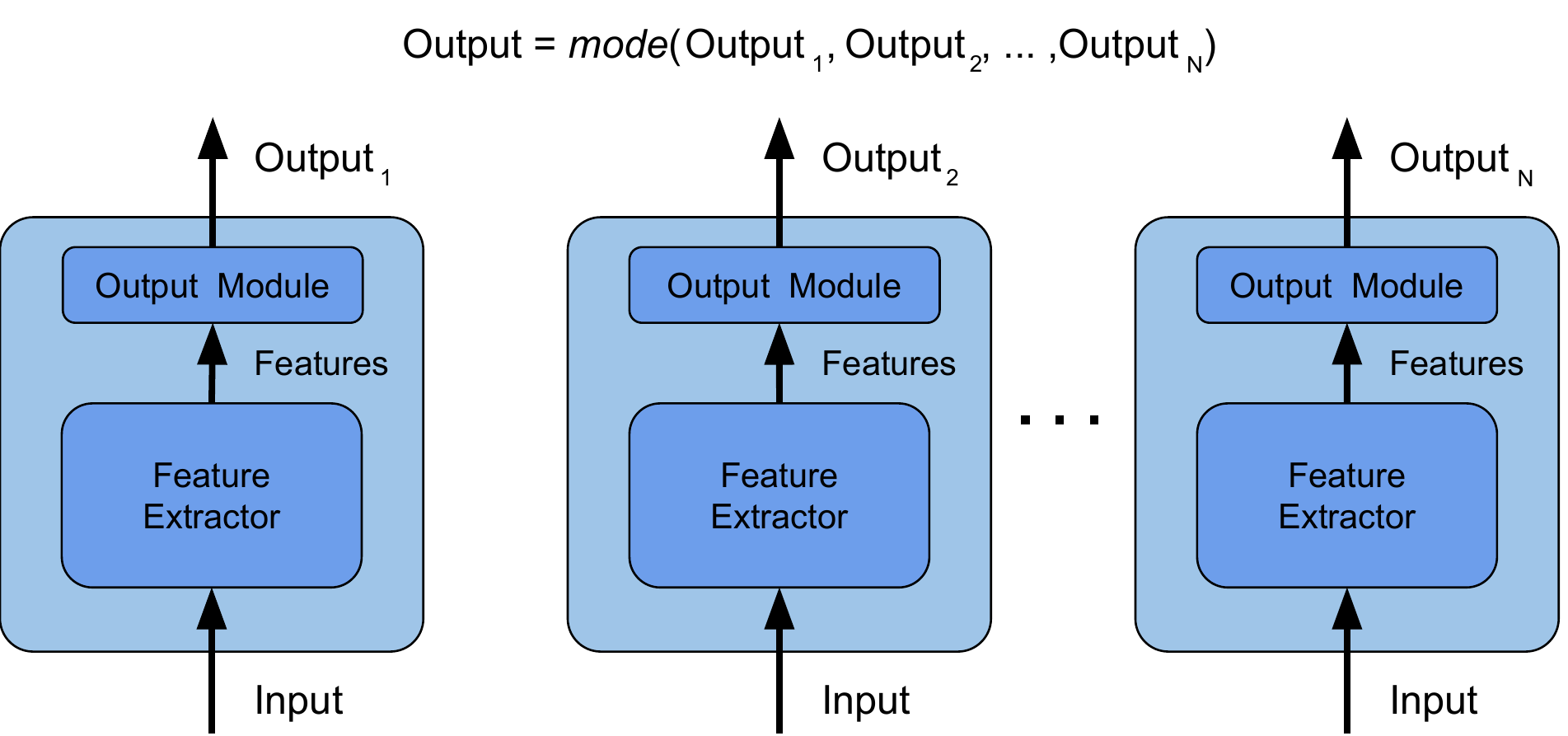}}
\caption{Ensemble by voting: the final output is obtained by picking the mode \rev{(i.e. most frequent class value)} among the output produced by the weak learners. In this way\rev{,} the weak learners are independent and voting is effective with a high number of heterogeneous weak learners.}
\label{fig:ensemble-voting}
\end{figure*}

\begin{figure*}[h]
\centering
\resizebox{.6\textwidth}{!}
{\includegraphics{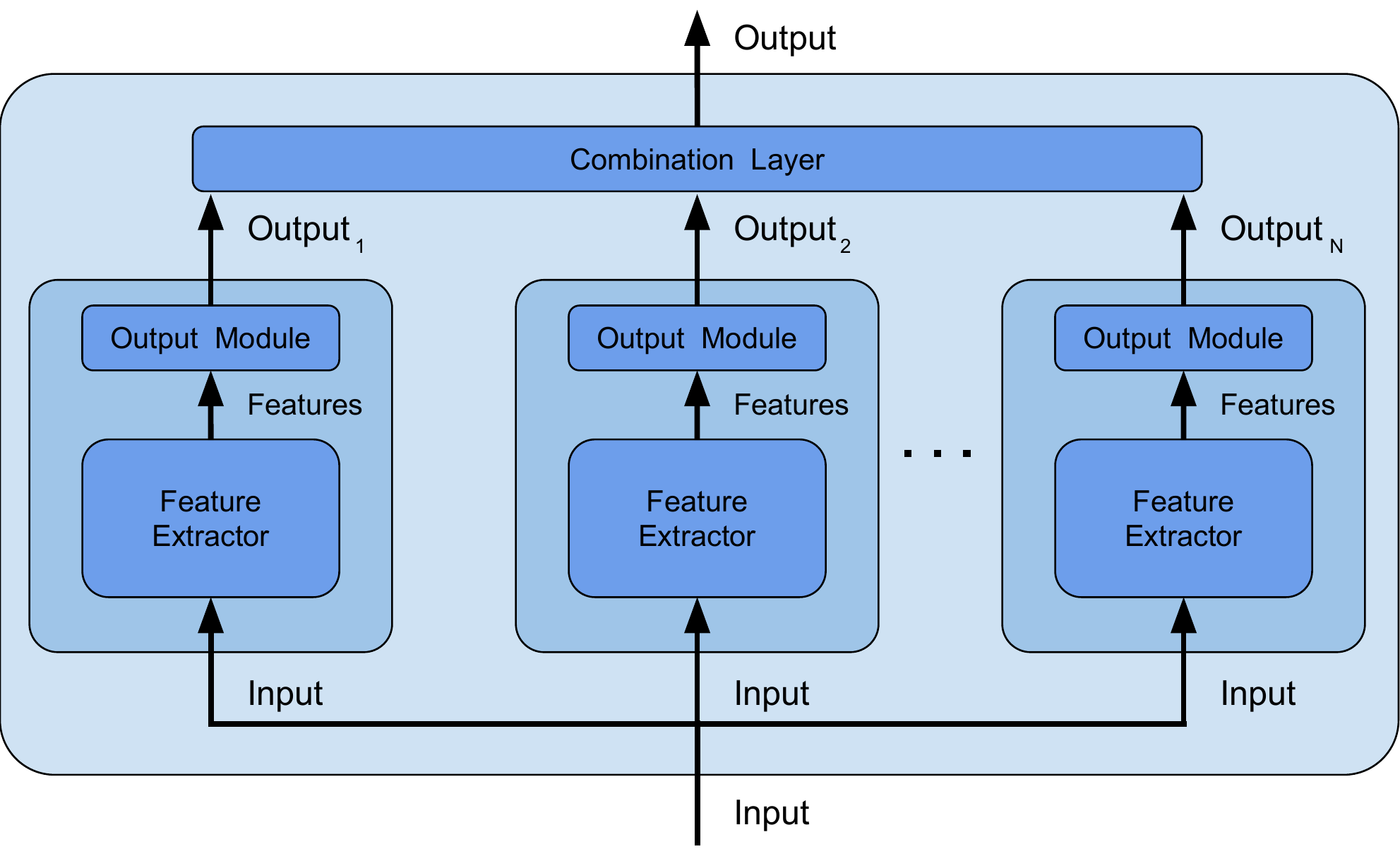}}
\caption{Ensemble by output combination: an additional combination layer is fed with the outputs of the weak learners and combines them. In this way, the weak learners are no longer independent and the combination layer can be trained to better adapt to data.}
\label{fig:ensemble-output}
\end{figure*}

\begin{figure*}[h]
\centering
\resizebox{.5\textwidth}{!}
{\includegraphics{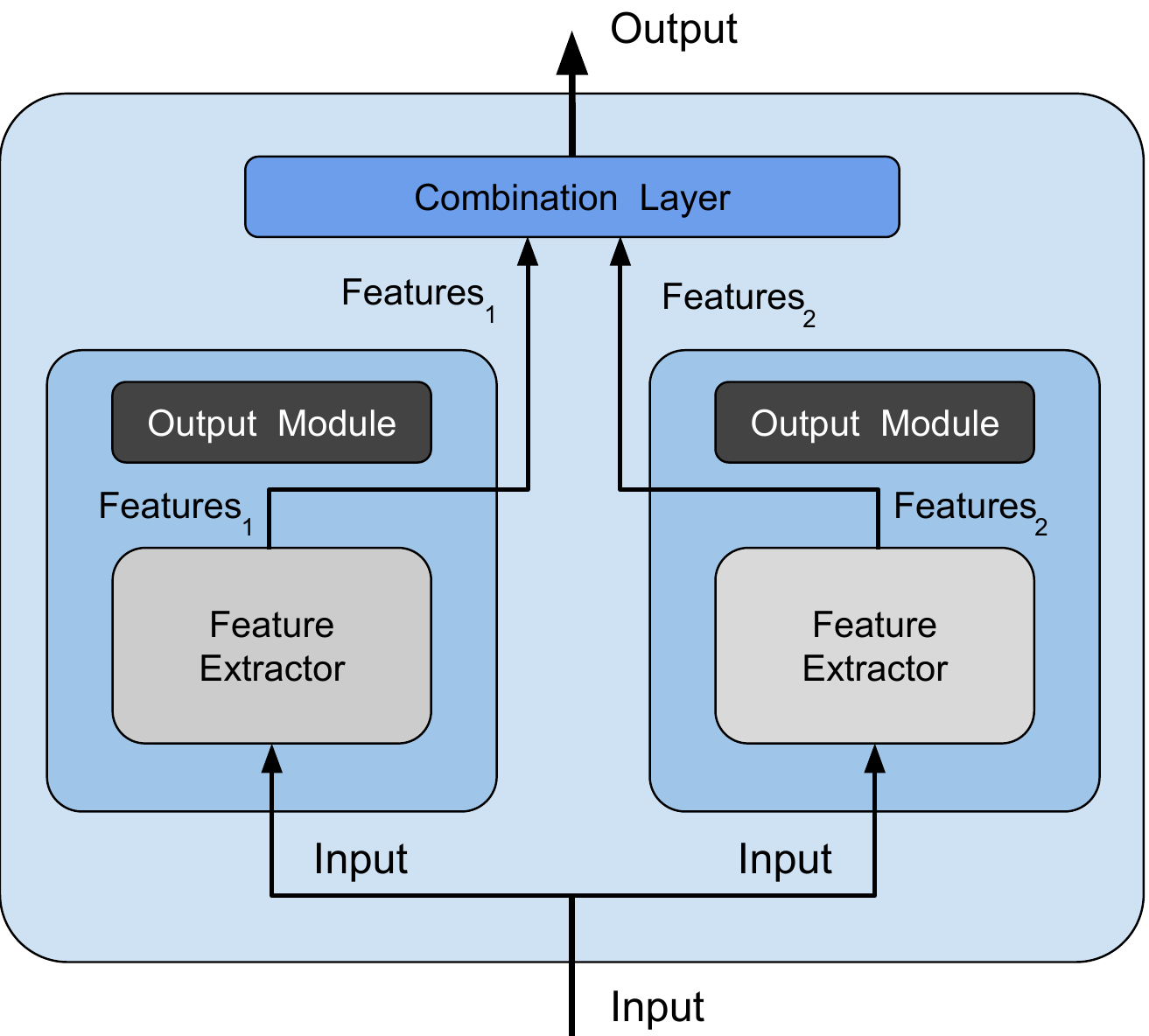}}
\caption{Our adaptive ensemble method: is an optimised version of the method shown in Figure \ref{fig:ensemble-output} because we avoid redundancy and reduce complexity by deleting the output module (dark grey-filled) of weak learners and feeding the combination layer with the features. Light grey-filled modules denote modules whose parameters are frozen during training. \rev{The diagram depicts the case $N=2$, which is used in most of the experiments in this paper, but the method can be applied with an arbitrary value for $N$.}}
\label{fig:ensemble-our}
\end{figure*}

The adaptivity is given by the fact that the \rev{proposed} ensembling is data-driven and not fixed as usual. The typical way of combining weak learners is to perform voting/averaging as shown in Figure \ref{fig:ensemble-voting} (predicting the output from all weak learners and then picking the most frequent output/average of \del{outputs}\rev{them}), respectively for classification/regression. However, in this case, the ensemble is only a static aggregator. In this work, we opted for performing an adaptive combination. However, instead of combining the outputs (Figure \ref{fig:ensemble-output}) of the weak learners, we combine the features that the CNNs extract from the input (\rev{see} Figure \ref{fig:ensemble-our} \rev{where the case $N=2$ is reported}). \rev{More formally, let $\mathrm{Feat}_{\mathrm{weak}_i}$ be the feature vector provided by feature extractor of the $i$-th weak learner and}

\begin{equation}
\mathrm{Feat}_{\mathrm{concat}}=\mathrm{Feat}_{\mathrm{weak}_1} \oplus \mathrm{Feat}_{\mathrm{weak}_2}\oplus \ldots \oplus \mathrm{Feat}_{\mathrm{weak}_{N-1}}\oplus \mathrm{Feat}_{\mathrm{weak}_N}
\end{equation}

\rev{be the vector contained by their concatenation. Then, the final fully connected final layer acts on the combined feature vector $\mathrm{Feat}_{\mathrm{comb}}$ defined as:}
\rev{
\begin{equation}
\mathrm{Feat}_{\mathrm{comb}}=W \cdot \mathrm{Feat}_{\mathrm{concat}}+b\end{equation}
}

In this way, we further reduce the complexity of the ensemble without reducing its power and expressiveness. Indeed\rev{,} the combination layer is of the same type \del{of} \rev{as} the output layer of the weak learners (i.e. Linear + LogSoftmax), and keeping both would introduce redundancy. This can be seen as a fully-differentiable version of Gradient Boosting \cite{gradient_boosting}. However, in this way, there is no reason to perform the tree decision traversal\rev{,} and the ensemble is performed at the features \del{-} level.

\section{Experimental results}\label{sec:results}
In this section, the results obtained on several major benchmark datasets on image classification are described. Before showing the results, the main aspects of the experimental setup are detailed. The experiments have been implemented using the PyTorch \cite{NEURIPS2019_9015} open-source machine learning framework.

\subsection{Datasets}
The proposed solution has been tested on several datasets in order to evaluate its capability of being effective over disparate domains (e.g. type of images, number of classes, balancing, quality) as shown in Table \ref{tab:datasetinfo}.
A brief description of each dataset follows:\\

\noindent{\bf CIFAR-10 and CIFAR-100 \cite{CIFAR}:} the CIFAR-10 dataset consists of 60000 32$\times$32 colour images in 10 classes, with 6000 images per class. There are 50000 training images and 10000 test images. CIFAR-100 dataset is just like the CIFAR-10, except it has 100 classes containing 600 images each. There are 500 training images and 100 testing images per class. The 100 classes in the CIFAR-100 are grouped into 20 superclasses. Each image comes with a ``fine'' label (the class to which it belongs) and a ``coarse'' label (the superclass to which it belongs). In the experiments, the fine-grained version with 100 classes has been used.\\

\noindent{\bf Stanford Cars \cite{stanfordcars}:} the Stanford Cars dataset contains 16185 360$\times$240 colour images of 196 classes of cars at the level of \emph{Make}, \emph{Model}, \emph{Year} (e.g. Tesla, Model S, 2012). The data is split into 8144 training images and 8041 testing images, where each class has been divided roughly in a 50-50 split. Since now, this dataset is referred as ``Cars''.\\

\noindent{\bf Food-101 \cite{food101}:} the Food-101 dataset consists of 101 food categories with 750 training and 250 test manually-reviewed images per category, making a total of 101000 images. On purpose, the training images contain some amount of noise that comes mainly in the form of intense colours and sometimes wrong labels. All images were rescaled to have a maximum side length of 512 pixels. \\

\noindent{\bf Oxford 102 Flower \cite{flower102}:} the Oxford 102 Flower is an image classification dataset consisting of 102 flower categories, most of them being plants commonly occurring in the United Kingdom. Each class consists of between 40 and 258 images. The images have large scale, pose and light variations. In addition, there are categories that have significant variations within the category and several very similar ones. Since now, this dataset is referred as ``Flower102''.\\

\noindent{\bf CINIC-10 \cite{CINIC10}:} 
CINIC-10 is a dataset for image classification consisting of 270000  32$\times$32 colour images. It was compiled as a ``bridge'' between CIFAR-10 and ImageNet, taking 60000 images from the former and 210000 downsampled images from the latter. It is split into three equal subsets - train, validation, and test - each containing 90000 images.\\

\noindent{\bf Oxford-IIIT Pet \cite{pets}:} the Oxford-IIIT Pet Dataset has 37 categories with roughly 200 images for each class representing dogs or cats (25 classes for dogs and 12 for cats). Different versions of the dataset can be used for image classification, object detection, or image segmentation. In particular, for the experimentation, the fine-grained version of the image classification task has been used (i.e. predict the particular breed of the animal in the image instead of just determining if it is a dog or a cat). The images have wide variations in scale, pose and lighting. Since now, this dataset is referred as ``Pets''.
\\
\begin{table*}[h]
\caption{Details about the datasets used in the  experiments.}
\newcommand\T{\rule{0pt}{2.5ex}}
\begin{center}
\begin{tabular}{lccccc}
    {\bf Dataset} &  {\bf Domain} & {\bf Input size} & {\bf Classes}  & {\bf Balanced} & {\bf Provided splits}\\
    \hline
    \T
    CIFAR-10 & Mixed (RGB) & 32$\times$32 & 10 & Yes & Train-Test \\
    CIFAR-100 & Mixed (RGB) & 32$\times$32 & 100 & Yes & Train-Test \\
    Cars & Cars (RGB) & 360$\times$240 & 196 & Yes & Train-Test \\
    Food-101 & Food (RGB) & 512 larger side & 101 & No & Train-Test \\
    Flower102 & Flowers (RGB) & Various & 102 & Yes & Train-Valid-Test \\
    CINIC-10 & Mixed (RGB) & 32$\times$32 & 10 & Yes & Train-Valid-Test \\
    Pets & Dogs \& Cats (RGB) & Various & 37 & Yes & Train-Valid-Test 
\end{tabular}
\end{center}
\label{tab:datasetinfo}
\end{table*}

\subsection{Input preprocessing}
The models are not fed directly with the images provided by the datasets, but images are preprocessed to improve the performances. In particular, the only two preprocessing steps done are resize (size chosen after preliminary tests) and stanmdardization (in order to have all data of the same dataset described under the same distribution with pixel values centred around the mean and unit deviation) which improves stability and convergence of the training. Preprocessing details for each dataset are shown in Table \ref{tab:preprocess}. Even if augmentation has been performed in the works reported as SOTA, no augmentation is performed in this work to test the performances of the ``pure'' method.

\begin{table*}[h]
\caption{Input sizes and standardization values, for each channel, used for data preprocessing.}
\newcommand\T{\rule{0pt}{2.5ex}}
\begin{center}
\begin{tabular}{lccc}
    {\bf Dataset} &  {\bf Input size} & {\bf Means (R,G,B)}  & {\bf Stds (R,G,B)}\\
    \hline
    \T
    CIFAR-10 & 256$\times$256 & (0.491, 0.482, 0.447) & (0.246, 0.243, 0.261)\\
    CIFAR-100& 256$\times$256 & (0.507, 0.486, 0.441) & (0.267, 0.256, 0.276)\\
    Cars & 500$\times$500 & (0.468, 0.457, 0.450) & (0.295, 0.294, 0.302)\\
    Food-101 & 500$\times$500 & (0.550, 0.445, 0.344) & (0.271, 0.275, 0.279)\\
    Flower102 & 500$\times$500 & (0.433, 0.375, 0.285) & (0.296, 0.245, 0.269)\\
    CINIC-10 & 256$\times$256 & (0.478, 0.472, 0.430) & (0.242, 0.238, 0.258)\\
    Pets & 500$\times$500 & (0.481, 0.448, 0.394) & (0.269, 0.264, 0.272)
\end{tabular}
\end{center}
\label{tab:preprocess}
\end{table*}

\subsection{Transfer learning}\label{transfer}
Transfer learning \cite{transfer} is the technique of taking knowledge gained while solving one problem,  and applying it to a different but related problem. Like most cases for image classification, the stored knowledge is brought by pre-trained models from the ImageNet \cite{imagenet} task,  since it has more than 14 million images belonging to 1000 generic classes. Transfer learning has been used for weak learners training only.

\subsection{Validation phases} \label{validation_phase}
Validation is divided into 2 main phases: \rev{end-to-end weak learner overfitting training and ensemble combination layer fine-tuning.}
\rev{In the first phase the transfer learning starts from the ImageNet pre-trained model and sets a new output module (to fit the output size). The models are trained to reach overfit in order to get high specialization on the subset they are referred to.
In the second phase, as shown in Figure \ref{fig:ensemble-our} the weak learners are frozen removing their output modules, so in this phase only the combination layer is trained.}

Both phases are performed using \rev{the}AdaBelief \cite{adabelief} optimizer which guarantees both fast convergence and generalization. AdaBelief parameters used are the following ones: learning rate $5\cdot10^{-4}$, betas (0.9, 0.999), eps $10^{-16}$, using weight decoupling without rectify.

\subsection{Avoid overfitting}\label{overfitting}
In order to prevent overfitting (i.e. avoid the model being too specialized to data from the training set with poor performances on \emph{unknown} data), we use early stopping  (i.e. stop training after no improvements on the validation set after a certain number of epochs, called \emph{patience}) during ensemble fine-tuning only.

\subsection{Data Splitting}
\label{sec:split}
Every dataset is provided with the ``official'' train-test split that is used for the ensemble fine-tuning. On the other hand, for the end-to-end overfitting training of the weak learners\rev{,} we perform the following data split:

\begin{enumerate}
\item set the size $N$ of the final ensemble model (i.e. the number of weak learners to \rev{be used in the} ensembl\rev{ing})\rev{:} in particular for the experiments $N=2$ in order to have the minimum ensemble size;
\item randomly split the training set into $N$ equally sized and disjoint (i.e. each data belongs exactly only to $1$ subset) subsets with stratification (i.e. preserving the class ratios within the subset). During the test \rev{only an}  exception \del{h}\rev{w}as made for the Pets dataset, in which the $2$ disjoint subsets were made only by cats and dogs, respectively;
\item for each subset, instantiate a weak learner and train it only on that subset (called \emph{bagging}), with overfitting. In this way every weak learner will be highly specialized only on that portion of data\rev{;} this could sound self-defeating but \cite{ensemble_overfit} \del{showed}\rev{has shown} that it leads to \rev{a} qualitative ensemble, especially in the case of this work in which ensembling is adaptive. The choice to reach overfitting will reduce the overall validation time: \del{with}\rev{ on the basis of}  preliminary tests, we noticed that EfficientNet-b0 and AdaBelief optimizer with overfitting training are powerful and will always converge to the same minimum point (very likely to be the global one, due to the fact that accuracy is 100\% almost always) independently on the initialization. In this way, just 2 train runs (only one initialization for each weak learner) are sufficient for every dataset.
\end{enumerate}

\subsection{Loss and Metrics}
\noindent{\bf Training Loss:} due to the multiclass nature of all dataset tasks, the Cross-Entropy Loss (which exponentially penalizes differences between predicted and true values, expressed as the probability of class belonging) is used. For this reason, the model output has a specified size depending on the dataset (i.e. the number of classes) and each element $output[i]$ represents the probability that the input sample belongs to class $i$.

\vspace{3mm}
\noindent{\bf Validation and test metrics:} for the validation set evaluation, we decided to use the Weighted F1-score because this takes into account both correct and wrong predictions (true/false positive/negative) and weighting allows to manage any imbalance of the classes (more representative classes have a greater contribution).
On the other hand, to make comparisons with previous works on the test set, we used the same metric, which is Accuracy (i.e. correct prediction/total set) in all cases.

\subsection{Hyperparameters}\label{hyperparameters}
Some hyperparameters have already been fixed \rev{and provided} in the previous sections (i.e. preprocessing size and standardization, optimizer parameters and ensemble size)\del{, to}\rev{.To} further reduce the total validation time other hyperparameters have been fixed:
\rev{early stopping patience was set to 10 epochs, batch size to 55 (200 in the case of fine-tuning) and 200 (700 in the case of the fine-tuning) for the 500$\times$500 and 256$\times$256 images, respectively.
}

\rev{
Here follows the hyperparameters configuration file for training a weak model, the same format is used for the ensemble, with the only difference that the ensemble\_module\_list parameter is not empty but contains the local addresses of the two best weak models:
}

{
\begin{lstlisting}[style=yaml]
project: projects/cifar10   # it varies depending on the dataset
seed: 9999  # it changes for all the runs
# means and standard deviations used for normalization varies depending on the dataset
means: [0.4918687788500817, 0.4826539051649305, 0.44717727749693625] 
stds: [0.24697121432552785, 0.24338893940435022, 0.2615925905215076]
early_stopping_patience: 10  
num_epochs: 100 # the maximum number of epoch (never reached)
image_size: 256 # size of the images depending on the dataset
batch_size: 200 
optim: AdaBelief  # the optimizer used
lr: 5e-4  # optimizer parameter
eps: 1e-16 # optimizer parameter
validation_metric:  F1 # F1-score is used as validation metric
from_pretrained: True # EfficientNet-b0 pretrained model from ImageNet is used
modeltype: efficientnet-b0
train_ratio: 0.8
valid_ratio: None   # automatically obtained 
test_ratio: None  # automatically obtained 
ensemble_module_list:  # in case of the ensemble it contains the local addresses of the weak models
\end{lstlisting}}

\rev{
As written before, there is no hyperparameters tuning, they are all prefixed except for the seeds}

\begin{itemize}
    \item \del{early stopping patience: 10 epochs;}
    \item \del{batch size: 55 (200 in the case of fine-tuning) and 200 (700 in the case of the fine-tuning) for the 500$\times$500 and 256$\times$256 images, respectively.}
\end{itemize}

For the ensemble fine-tuning, 5 different random seeds are used.
In this way, for each dataset, 2 end-to-end weak training (1 for each subset) and 5 fine-tuning ensemble training are performed.

\section{Results and Discussion}
In this section, the results of the experiments are shown and  discussed. Table \ref{tab:results} shows that our work improves the SOTA in all major benchmark datasets and as expected the highest improvements ($>$ 0.5\%) are obtained on the tasks which are not saturated (i.e. accuracy $<$ 99\%). These results gain more evidence when complexity is considered too: indeed Table \ref{tab:complexities} shows that our work (except \del{for}\rev{in} the case of CINIC-10) has 5-60 times less total number of parameters and needs 10-100 times fewer FLOPs respect to the SOTA. Moreover, in terms of trainable parameters, since it performs the fine-tuning of a combination layer, our final solution has only about 100K parameters to train.

\begin{table*}[h]
\caption{Classification test accuracy comparison between SOTA and our work on datasets used during experiments.}
\newcommand\T{\rule{0pt}{2.5ex}}
\begin{center}
\begin{tabular}{lccc}
    {\bf Dataset} &  {\bf SOTA accuracy} & {\bf Our accuracy}  & {\bf Improvement}\\
    \hline
    \T
    CIFAR-10 \cite{vith14} & 99.500\% & 99.612\% & 0.112\%\\
    CIFAR-100 \cite{SAM} & 96.080\% & 96.808\% & 0.728\%\\
    Cars \cite{tresentlv2} & 96.320\% & 96.868\% & 0.548\%\\
    Food-101 \cite{SAM} & 96.180\% & 96.879\% & 0.699\%\\
    Flower102 \cite{cvt24} & 99.720\% & 99.847\% & 0.127\%\\
    CINIC-10 \cite{NATm3} & 94.300\% & 95.064\% & 0.764\%\\
    Pets \cite{SAM} & 97.100\% & 98.220\% & 1.120\%
\end{tabular}
\end{center}
\label{tab:results}
\end{table*}

\begin{table*}[h]
\caption{Complexity, both number of parameters and FLOPs, comparison between SOTA and our work on datasets used during experiments.}
\begin{center}
\begin{threeparttable}
\newcommand\T{\rule{0pt}{2.5ex}}

\begin{tabular}{lccccc}
    {\bf Dataset} &  {\bf SOTA parameters} & {\bf Our parameters} & {\bf SOTA FLOPs} & {\bf Our FLOPs}\\
    \hline
    \T
    CIFAR-10 \cite{vith14} & $\approx$ 632M & $\approx$ 11M (100K) & $\approx$ 916G$^{\dagger}$& $\approx$ 0.9G\\
    CIFAR-100 \cite{SAM} & $\approx$ 480M & $\approx$ 11M (100K) & $\approx$ 299G$^{*}$ & $\approx$ 0.9G\\
    Cars \cite{tresentlv2} & $\approx$ 54.7M & $\approx$ 11M (100K) & $\approx$ 10G & $\approx$ 0.9G\\
    Food-101 \cite{SAM} & $\approx$ 480M & $\approx$ 11M (100K) & $\approx$ 299G$^{*}$& $\approx$ 0.9G\\
    Flower102 \cite{cvt24} & $\approx$ 277M & $\approx$ 11M (100K) &$\approx$ 60G& $\approx$ 0.9G\\
    CINIC-10 \cite{NATm3} & $\approx$ 8.1M & $\approx$ 11M (100K) & $\approx$ 1G & $\approx$ 0.9G\\
    Pets \cite{SAM} & $\approx$ 480M & $\approx$ 11M (100K) & $\approx$ 299G$^{*}$& $\approx$ 0.9G
\end{tabular}

\begin{tablenotes}
\small
\item{$\dagger$ Estimation based on a similar architecture with a similar number of parameters.}
\item{$*$ Estimation based on the same architecture but scaling FLOPs w.r.t. the number of parameters ratio.}
\end{tablenotes}
\end{threeparttable}
\label{tab:complexities}
\end{center}
\end{table*}

\rev{In order to stress our method, we also provide a different combination of weak classifiers: specifically, we show the results of an ensemble of five weak models. For demonstration purposes we report the results obtained only for the CIFAR-100 and CIFAR-10 datasets.
In the case of CIFAR-100, while the ensemble using 2 weak models obtained an accuracy of 96.808\%, the new one obtained an accuracy of 84.930\%. This result was expected, since  each weak model had to be trained on a third of the images of the previous case according to the data splitting procedure described in Section \ref{sec:split} in order to avoid the use of the same images.
In the case of CIFAR-10, while the ensemble using 2 weak models obtained an accuracy of 99.612\%, the new one obtained an accuracy of 96.640\%
}

\rev{Again both for CIFAR-10 and for CIFAR-100, we also run 6 EfficientNet-b0 weak models and then 6 classical ensembles by majority voting in order to further compare the classical method with ours: one ensemble collects the best two weak models and another the best five ones. For CIFAR-10, the best weak model reaches 97.37\% of accuracy, the best ensemble of 2 weak models reaches 97.54\% and the ensemble of 5 weak models 97.66\% (our method reaches 99.61\%).
For CIFAR-100 the best weak model reaches 85.55\% of accuracy, the ensemble of 2 weak models reaches 86.64\% and the best ensemble of 5 weak models 87.56\% (our method reaches 96.81\%).
Moreover, we also run our ensemble method on these classical weak models to show that, as we described in Section \ref{sec:our_work}, our solution improves the results both for the novelties applied to the weak models  and for those applied to the ensemble.
With CIFAR-10 the best adaptive ensemble reaches 97.49\% (against our full method that reaches 99.61\%) and with CIFAR-100 the best adaptive ensemble reaches 86.79\% (against our full method that reaches 96.81\%).
}

Last but not least, below we present an analysis of computation time for a single task; let:

\begin{itemize}
    \item $T_{\mathrm{end}}$ the time for a single end-to-end weak learner training;
    \item $T_{\mathrm{fine}}$ the time for a single fine-tuning ensemble model training;
    \item $T_{\mathrm{fwd}}$ the time for a single forward step;
    \item $T_{\mathrm{back}}$ the time for a single backward step, when subscripted it indicates the number of parameters involved;
    \item $T_{\mathrm{upd}}$ the time for a single optimization update step, when subscripted it indicates the number of parameters involved.
\end{itemize}

Then, for a single task, the total time needed is:

\begin{equation}
T = A \cdot T_{\mathrm{end}}+B \cdot T_{\mathrm{fine}}
\label{eq:time1}
\end{equation}

where in our case $A=2$ since end-to-end training is performed once on each of the two disjoint subsets and $B=5$ because we performed fine-tuning ensemble training with five random initializations.

However, it is possible to perform in parallel each of the end-to-end training processes, halving the batch size and about taking half of the time; the same goes for the fine-tuning training running all in parallel, in this way the total time is:

\begin{equation}
T = T_{\mathrm{end}}+T_{\mathrm{fine}}
  \label{eq:time2}
\end{equation}

and considering that a single training  is made of forward+backward+update steps to all training data for several epochs:

\begin{equation}
T_{end} \propto T_{\mathrm{fwd}} + T_{\mathrm{back}} + T_{\mathrm{upd}} \\
  \label{eq:time3}
\end{equation}

\begin{equation}
T_{\mathrm{fine}} \propto \del{K} \rev{N} \cdot T_{\mathrm{fwd}} + T_{\mathrm{back}_{100k}} + T_{\mathrm{upd}_{100k}} \appropto T_{\mathrm{fwd}}
  \label{eq:time4}
\end{equation}

\rev{that is the time for a single fine-tuning ensemble model training is proportional (depending on the actual number of epochs) to the number of the weak learners N multiplied for the time needed for a single forward step, plus the time for a single backward step using 100000 parameters, plus the time for a single optimization update step using 100000 parameters that is approximately proportional to the time for a single forward step. Indeed, t}he approximation in the Equation \ref{eq:time4} is \del{done because} \rev{justified by the fact that} backward and update steps involve only a small fraction of the parameters; moreover, the two weak learners perform forward steps in parallel since they are independent (otherwise we should have $K=2$). Putting together Equations \ref{eq:time3} and \ref{eq:time4}, the total time is:

\begin{equation}
T \appropto 2 \cdot T_{\mathrm{fwd}} + T_{\mathrm{back}} + T_{\mathrm{upd}}
  \label{eq:time6}
\end{equation}
\rev{that is,  the total time is proportional to 2 multiplied the time for a single forward step plus the time for a single backward step plus the time for a single optimization update step.}

What said before, in terms of FLOPs is (considering only one input, just add the linear scaling factor for the training on the whole dataset):

\begin{equation}
F_{\mathrm{fwd}} = F_{\mathrm{back}} = 0.39 \ \text{GFLOPs} 
  \label{eq:effflops}
\end{equation}

\begin{equation}
F_{\mathrm{upd}} \approx 20 \cdot P \approx 0.1 \ \text{GFLOPs}
  \label{eq:adaflops}
\end{equation}

the Equation \ref{eq:effflops} refers to FLOPs of EfficientNet-b0 architectures and the Equation \ref{eq:adaflops} refers to FLOPs of AdaBelief update step where $P = 5$M is the number parameters involved in the end-to-end training. Putting all together:

\begin{equation}
\begin{gathered}
F \approx 2 \cdot F_{\mathrm{fwd}} + F_{\mathrm{back}} + F_{\mathrm{upd}} \approx \\
\approx 2 \cdot 0.39 + 0.39 + 0.1 \approx\\
\approx 1.3 \  \text{GFLOPs}
  \label{eq:flops}
\end{gathered}
\end{equation}

this means that the \emph{whole pipeline on a single image} requires about 1.3 GFLOPs, and considering the Table \ref{tab:complexities}, the SOTA for CINIC-10 in \cite{NATm3} that has the least number of parameters (8.1M) requires  1 GFLOPs for \emph{one single forward on an image}, showing that our solution is the fastest and the speedup is much more noticeable (10-100 times) over the even more complex SOTA models.

\section{Conclusion and future works}\label{conclusion}

In this work, we presented a method to reverse the trend in image classification of having minor improvements with a huge complexity increase. In particular, we showed a revisited \emph{ensembling} to outperform the SOTA with restrained complexity, both in terms of the number of parameters and FLOPs. Specifically, we proved how it is possible to perform bagging on two disjoint subsets of data using two EfficientNet-b0 weak learners and training them to overfit on the assigned/scheduled subset. 

In this work we pushed the ensemble size to the lower bound using only 2 weak learners: this adaptive ensemble strategy would still be the most efficient using up to 5 weak learners \rev{(taking into account that, when using the overfitting strategy, each weak learner has too fragmented and limited knowledge)}, and then it could be further improved by defining different bagging strategies (e.g. train weak learners on subsets split by class dimensionality, clustering or different color space mapping of inputs).

Then, the ensemble is performed by fine-tuning a trainable combination layer. The efficiency of the method is given by different reasons: efficiency of EfficientNet-b0 models, fine-tuning for ensemble and the high parallelization capability of the solution, the reduced number of FLOPs combined with the tiny validation space (7 total runs: 2 end-to-end + 5 fine-tuning).

These results pave to investigate this kind of strategy in many fields: Object Detection (performing the ensemble at feature extraction backbone level) and Segmentation (performing the ensemble on the encoding in typical encoder-decoder architectures).

\bibliography{biblio}

\end{document}